%% file: main.tex
\RequirePackage{fix-cm}
\documentclass[smallcondensed]{svjour3}

\input{includepackages}
\usepackage[square,comma,numbers]{natbib}

\begin{document}
\nocite{*}
\title{Reinforcement Learning in Non-Stationary Environments}
\author{Sindhu Padakandla         \and
        Prabuchandran K. J \and
        Shalabh Bhatnagar}
        
\institute{Sindhu Padakandla \at
              Dept. of Computer Science and Automation, Indian Institute of Science, Bangalore \\
              \email{sindhupr@iisc.ac.in}           
           \and
           Prabuchandran K. J \at
            Dept. of Computer Science and Automation, Indian Institute of Science, Bangalore \\
            \email{leeaxil@gmail.com} \\
            \emph{Dept. of Computer Science and Engineering, Indian Institute of Technology, Dharwad}
            \and 
            Shalabh Bhatnagar \at
            Dept. of Computer Science and Automation, Indian Institute of Science, Bangalore\\
            \email{shalabh@iisc.ac.in}
}


\date{}
\begin{abstract}
Reinforcement learning (RL) methods learn optimal decisions in the presence of a stationary environment. However, the stationary assumption on the environment is very restrictive. In many real world problems like traffic signal control, robotic applications, etc., one often encounters situations with non-stationary environments and in these scenarios, RL methods yield sub-optimal decisions. In this paper, we thus consider the problem of developing RL methods that obtain optimal decisions in a non-stationary environment. The goal of this problem is to maximize the long-term discounted reward accrued when the underlying model of the environment changes over time. To achieve this, we first adapt a change point algorithm to detect change in the statistics of  the environment and then develop an RL algorithm that maximizes the long-run reward accrued. We illustrate that our change point method detects change in the model of the environment effectively and thus facilitates the RL algorithm in maximizing the long-run reward. We further validate the effectiveness of the proposed solution on non-stationary random Markov decision processes, a sensor energy management problem and a traffic signal control problem.

\end{abstract}
\maketitle

\section{Introduction}
\label{sec:intro}
\input{intro-ai2019}

\section{Related Work}
\label{sec:rw}
\input{relatedwork}

\section{Preliminaries}
\label{sec:prelim}
\input{prelim-ai2019}

\section{Problem Formulation}
\label{sec:pf}
\input{pf-ai2019}

\section{Our Approach}
\label{sec:method}
\input{method-ai2019}
\section{Experimental Results}
\label{sec:expres}
\input{expres-ai2019}

\section{Conclusions}
\label{sec:fw}
This work develops a model-free RL method known as Context QL for learning optimal policies in changing environment models. 
A novel change detection algorithm for experience tuples is used to determine changes in environment models in conjunction with QL.
The numerical experiments in realistic applications show that Context QL is promising, since, the policies learnt by the method are seen to perform well in varying operating conditions and give better return, when compared to classical QL and other non-stationary learning algorithms like RUQL and RLCD. We additionally showed the statistical performance of Context QL with respect to the precision and recall metric, which is an important metric in time series and changepoint literature.

Future enhancements to this work can focus on detecting changes in the context of large and continuous state space models. Such an extension will indeed be useful in robotics and intelligent transportation systems. Also, if the sequence of context changes is not known, then we require some changes to the proposed method. An interesting extension involves \emph{meta-learning}~\cite{iclr2019} which can be adapted to continuous and large state-action space MDPs.

\bibliographystyle{spbasic}
\bibliography{springer_ai_jrnl_ref}

%
%
%
%
%
%
%




\end{document}

%% file: includepackages.tex
\usepackage[utf8]{inputenc}
\usepackage{amsmath}
\usepackage{amsfonts,amssymb}
\usepackage{url}
\usepackage{graphicx}

\usepackage{float}
\usepackage{multirow}

\usepackage{tikz}
\usepackage{pgf}
\usepackage{placeins}

\usepackage{comment}
\usepackage{url}
\usepackage{placeins}
\usepackage{latexsym}
\usepackage{pifont}
\usepackage{pgfplots,filecontents}
\usepackage{gensymb}
\usepackage{url}
\usepackage{caption}
\usepackage[symbol]{footmisc}
\usepackage{algorithm}
\usepackage{algorithmic}
\usepackage{fullpage}

\pgfplotsset{compat=newest}

\usetikzlibrary{intersections}
\usetikzlibrary{positioning,angles,quotes,patterns}

\newcommand{\R}{\mathbb{R}}

\newcommand{\E}{\mathbb{E}}
\newcommand{\ra}{\rightarrow}

\newtheorem{assumption}{Assumption}

\DeclareMathOperator*{\argmax}{arg\,max}

\DeclareMathOperator*{\sign}{sign}

\DeclareGraphicsExtensions{.png,.jpg,.jpeg,.eps,.gif}
\tikzstyle{block} = [rectangle, draw,text width=3em, text centered, rounded corners]

\tikzstyle{block2} = [rectangle, draw, text centered, rounded corners]
    
\tikzstyle{line} = [draw, -latex]

%% file: intro-ai2019.tex
Autonomous agents are increasingly being designed for sequential decision-making tasks under uncertainty in various domains. For e.g., in traffic signal control \cite{prashla-tsc}, 
an autonomous agent decides on the green signal duration for all lanes at a traffic junction, while in robotic applications, human-like robotic agents are built to dexterously manipulate 
physical objects \cite{openai2018learning,robotics2}. The common aspect in these applications is the evolution of the \emph{state} of the system based on decisions by the agent. In traffic signal control for instance, the \emph{state} is the vector of current congestion levels at the various lanes of a junction and the agent decides on the green signal duration for all lanes at the junction, while in a robotic application, the \emph{state} can be motor angles of the joints etc., and the robot decides on the torque for all motors. 
The key aspect is that the decision by the agent affects the immediate next state of the system, the \emph{reward (or cost)} obtained as well as the future states. 
Further, the sequence of decisions by the agent is ranked based on a fixed \emph{performance criterion}, which is a function of the rewards obtained for all decisions made. The central problem in \emph{sequential decision-making} is that the agent must find a sequence of decisions for every state such that this performance criterion is optimized.
Markov decision processes, dynamic programming (DP) and reinforcement learning (RL) \cite{BertB,puterman,sutton} provide a rich mathematical framework and algorithms which aid an agent in sequential decision making under uncertainty.

\begin{figure}[h!]
\begin{center}
\begin{tikzpicture}[node distance = 14em, auto, thick]
    \node [block] (Agent) {Agent};
    \node [block2, below of=Agent] (Environment) {\includegraphics[scale=0.25]{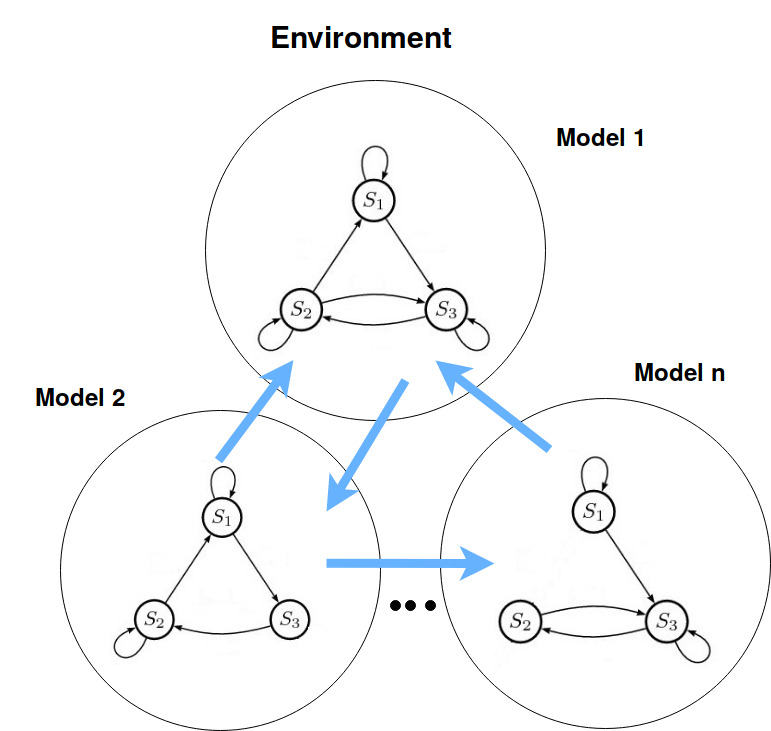}};
    
    \draw[->] (Agent) -- node [midway,,align=center]{Action\\ $a_t$} (Environment);
    \draw[->] (Environment.east) --++(0.8,0) --++(0,4.2) -- node [midway,above,align=center] {Next State $s_{t+1}$}(Agent.east);
    \draw[->] (Environment.west) --++(-0.5,0) --++(0,4.2) -- node [midway,above,align=center]{Reward $r_{t}$} (Agent.west);
\end{tikzpicture}
\end{center}
\caption{Non-Stationary RL Framework} \label{fig:rl}
\end{figure}
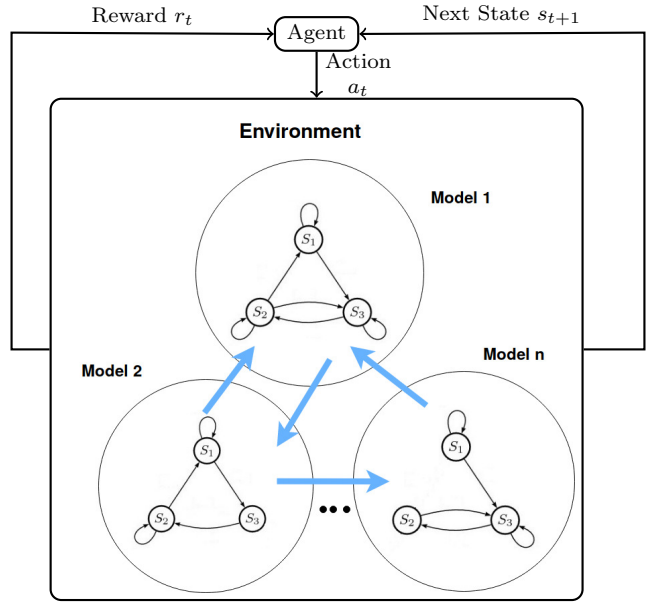

In this paper, we consider an important facet of real-life applications where the agent has to deal with non-stationary rewards and non-stationary transition probabilities between system states. 
For example, in vehicular traffic signal control, the traffic inflow rate in some (or all) lanes is quite different during peak and off-peak hours. The varying traffic inflow rates makes some lane queue length configurations more probable compared to other configurations, depending 
on the peak and off-peak traffic patterns.
It is paramount that under such conditions, the agent selects appropriate green signal duration taking into account the different traffic patterns.
Also, in robotic navigation \cite{robotics2}, the controller might have to vary robotic arm/limb joint angles depending on the terrain or weather conditions to ensure proper locomotion, because the same joint angles may give rise to different movement trajectories in varying terrains and weather conditions.
When environment dynamics or rewards change with time, the agent must quickly adapt its policy to maximize the long-term cumulative rewards collected and ensure proper and efficient system operation as well. We view this scenario as illustrated in Fig. \ref{fig:rl}, where the environment changes between models $1,2,\ldots,n$ dynamically. 
The epochs at which these changes take place are unknown to (or hidden from) the agent controlling the system. 
The implication of the non-stationary environment is this: when the agent exercises a control $a_t$ at time $t$, the next state $s_{t+1}$ as well as the reward $r_t$ are functions of the \emph{active} environment model dynamics.

Motivated by the real-world applications where changing environment dynamics (and/or rewards, costs) is frequently observed, we focus 
on developing a model-free RL method that learns optimal policies for non-stationary environments.

\subsection{Our Contributions} 
\begin{itemize} 
\item The primary contribution of this paper is to propose a model-free RL algorithm for handling non-stationary environments.
In this work, we adapt Q-learning (QL) \cite{ql} to learn optimal policies for different environment models.
\item The new method is known as Context Q-learning (see Section \ref{subsec:cql}). It is a \emph{continual learning}~\cite{crl1} algorithm that is designed for dynamically changing environments. 
\item Context Q-learning utilizes data samples collected during learning to detect changes in the model.
To achieve this, it leverages a novel change detection algorithm~\cite{prabukj} for these samples. Using results of change detection, the method estimates policy for the new model or improves the policy learnt, if the model had been previously experienced. In this manner, our method avoids catastrophic forgetting~\cite{cf}.
\item Context Q-learning is an online method which can learn and store the policies for the different environment contexts.
We assume that model-change patterns are known. Our method is novel in the sense that we utilize state and reward samples to detect these changes, unlike previous works which need model information. Thus, our method is model-free and can be used in real-world applications of RL. 
\item The experimental results (see Section \ref{sec:expres}) show the performance of Context QL on standard problems as well as on interesting problems in traffic signal control and sensor networks. The performance is evaluated on metrics like mean detection delay as well as precision and recall. This is in addition to the main performance metric, which is the reward collected by the algorithm in presence of dynamic environments. The results are compared with recent relevant works and the advantages of Context Q-learning are highlighted.
\end{itemize}
\subsection{Organization of the Paper}
The rest of the paper is organised as follows. In the following Section, we discuss prior approaches to this problem. In Section \ref{sec:prelim}, we give a brief background of Markov decision process (MDP) framework. 
This section describes the basic definitions and assumptions made by the DP algorithms for solving MDPs. 
Section \ref{sec:pf} describes the problem in mathematical terms. Additionally, it also introduces the notation that will be used in the rest of the paper.
We propose an RL method for non-stationary environments in Section \ref{sec:method}. 
Section \ref{sec:expres} shows numerical results on different application domains and analyzes the results. Finally, Section \ref{sec:fw} provides the concluding remarks.

%% file: relatedwork.tex
Very few prior works have considered the problem of developing RL algorithms for non-stationary environment models. 
\cite{choi1,choi2} proposed modeling changing environments in terms of
hidden-mode MDPs (HM-MDPs), wherein each \emph{mode (or context)} captures a stationary MDP setting and mode transitions are hidden.
All modes share the state and action spaces, but differ either in transition probability function of system states and/or reward function.
The methods described in \cite{choi2} require information about these functions for each of the modes.
Additionally, algorithms which find optimal policies for systems modeled as HM-MDP are computationally intensive and are not practically implementable. 

A context detection based RL algorithm (called RLCD) is proposed in \cite{rlcd}. The RLCD algorithm estimates transition probability and reward functions 
from simulation samples, while predictors are used to assess whether these underlying MDP functions have changed. 
The active context which could give rise to the current state-reward samples is selected based on an error score. 
The error score of all contexts  is estimated till the current epoch. The context which minimizes this error score is 
designated as the current active model. If all the contexts have a high error score, a new context is estimated.
RLCD does not require apriori knowledge about the number of environment contexts, but is highly memory intensive, since it 
stores and updates estimates of transition probabilities and rewards corresponding to all detected contexts. 
Moreover, the predictor which allows detection of new contexts is heuristic and is not easy to interpret.

Theoretical framework for RL in fast changing environments based on $(\epsilon,\delta)$-MDP is developed in \cite{csaji}. 
In this framework, if the accumulated changes in transition probability or reward function remain \emph{bounded} 
over time and are insignificant, then \cite{csaji} shows that changes in the optimal value function
are also negligible. However, this work does not provide a control algorithm which adapts to  the
changes in environment. 

Regret minimization algorithms proposed in \cite{cmdp,ucrl2,gajane2019paper1} study MDPs with varying transition probability and reward functions. 
These works consider a finite horizon $T$ and minimize the regret over this horizon, when the environment changes utmost $K$ times 
(in \cite{cmdp,ucrl2}) or arbitrarily (in \cite{gajane2019paper1}).
In both cases, the objective of algorithm proposed in \cite{cmdp}, the UCRL2 algorithm \cite{ucrl2} and variation-aware UCRL2 \cite{gajane2019paper1} is to reduce the sum of missed rewards compared to the rewards yielded by optimal policies in the periods during which the environment parameters remain constant. 
However, the optimal policies defined in these works differ with respect to the performance criterion.
In \cite{ucrl2}, the optimal policies are stationary and average-reward optimal, while \cite{cmdp,gajane2019paper1} considers an optimal (possibly non-stationary)   optimal policy to be total-reward optimal. \cite{cmdp} considers a non-stationary optimal policy, which is time-ordered using as components total-reward optimal policies of all contexts.

\cite{cmdp} considers the setting wherein the time horizon $T$ is divided into $H$ episodes, with an MDP context picked at the start of each episode. 
After the context is picked (probably by an adversary), a start state for the episode is also selected. The context is selected from a finite set $\mathcal{C}$, ($|\mathcal{C}| = K$) but is hidden from the RL controller. The algorithm clusters the observed episodes into one of the $K \leq T-1$  models and classifies an episode as belonging to one of these clusters. Depending on the cluster chosen, the context is explored and rewards are obtained.


The UCRL2 and variation-aware UCRL2 algorithms estimate the transition probability function as well as the reward function for an environment, and when the environment changes, the estimation restarts leading to a loss in the information
collected. The objective of these algorithms is to minimize the regret during learning and not to find the appropriate policies for the different environment settings. Hence if the environment pattern alternates between two different settings (say A and B), i.e., if the change pattern is A-B-A-B, then these algorithms restart estimation of transition probability and reward functions from scratch for both environments A and B when the second time these environments are encountered. In contrast, the objective of our method is to learn appropriate policies for each of the environments without discarding the previously gathered information.


A model-based method for detecting changes in environment models was proposed in \cite{taposh1}, while \cite{hadoux} proposes an extension to the RLCD method. Both these works employ quickest change detection \cite{shiryaev} methods to detect changes in the transition probability function and /or reward function.
The approach in \cite{hadoux} executes the optimal policy for each MDP, while parallely using cumulative sum (CUSUM)~\cite{page} technique to find changes. \cite{taposh1} shows that such an approach leads to loss in performance with delayed detection. It designs a two-threshold switching policy based on Kullback-Leibler (KL) divergence that detects changes faster, although with a slight loss in rewards accrued. However, \cite{taposh1} is limited in scope, since it assumes that complete model information of all the contexts is known. 
Hence, the work is not applicable in model-free RL settings. 
Moreover, \cite{taposh1} does not specify any technique for selecting the threshold values used in the switching strategy, even though the method proposed is completely reliant on the threshold values chosen.

A variant of Q-learning (QL), called the Repeated Update QL (RUQL) was proposed in \cite{ruql}. It essentially repeats the updates to the Q values of a state-action pair and is shown to have learning dynamics which is better suited to non-stationary environment tasks, on simulation experiments.
However, the RUQL faces the same issues as QL - it can learn optimal policies for only one environment model at a time.
QL and RUQL update the same set of Q values, even if environment model changes.
Additionally, unlike \cite{rlcd,taposh1,hadoux}, QL and RUQL do not incorporate mechanisms for monitoring changes in environment.
So, the RL agent utilizing QL or RUQL to learn policies cannot infer whether the model has changed. This leads to the scenario that when the environment parameters change, the RL agent starts learning policy for the new environment by completely discarding the policy learnt using samples obtained from the environment before the change.

A related problem is \emph{concept drift detection}~\cite{harel2014concept,concept_drift_survey,rlconceptdrift,tnnls}, which has been well-studied in supervised learning~\cite{harel2014concept}, semi-supervised learning and data mining~\cite{concept_drift_survey}. This area too addresses the problem of learning in dynamic environments. In supervised and semi-supervised learning, all instances of the training data are usually assumed to be generated from the same ``source" or ``concept" (though samples may be noisy). However, in domains like recommendation systems, bio-signals, concept drift occurs, wherein the same labelled dataset has different sources. This is very similar to the problem we address in this work. However, unlike the supervised and semi-supervised learning approaches, concept drift in RL environment has not been explored in the literature. A recent work~\cite{rlconceptdrift} formulates agent model retraining in presence of concept drift as a RL problem.~\cite{tnnls} is a recent work on concept drift in Markov chains. The authors propose concept drift detection mechanisms which is limited to finding the change in transition probabilities of Markov chains. However, this work can be used for concept drift in model-based RL, as highlighted in Section \ref{sec:expres}.

While all the prior works have provided significant insights into the problem, there are still issues with computational efficiency. Moreover, a model-free RL technique is needed which can retain past information and utilize it to learn better policies for all observed contexts. In the next section, we describe the underlying mathematical formulation for RL and the assumptions which are the basis of all classical RL algorithms.

%% file: prelim-ai2019.tex
A Markov decision process (MDP) \cite{BertB} is formally defined as a tuple $M = \langle S, A, P, R\rangle$, where $S$ is the set of states of the system, $A$ is the set of actions (or decisions) and $P:S \times A \times S \ra [0,1]$ is the transition probability function. The transition function $P$ models the uncertainty in the evolution of states of the system based on the action exercised by the agent. The evolution is uncertain in the sense that given the current state $s$ and the action $a$, the system evolves into the next state according to the probability distribution $P(s,a,\cdot)$ over the set $S$.  
Actions are selected at certain \emph{decision epochs} by the agent based on their \emph{feasibility} in the given state. A decision epoch is the time instant at which the agent selects an action and the number of such epochs determines the decision horizon of the agent.
When the number of decision epochs is infinite, we refer to $M$ as an \emph{infinite-horizon} MDP.
Depending on the application, each action yields a numerical reward (or cost), which is modeled by the function $R: S \times A \ra \R $. Transition function $P$ and reward function $R$ define the \emph{environment model} in which the system operates and the agent interacts with this environment. The interaction comprises of the action selection by the agent for the state and 
the environment presenting it with the future state and reward (or cost) for the action selected. 

A deterministic decision rule $d: S \ra A$ maps a state to its feasible actions and it models the agent's action choice for every state. The agent picks a decision rule for every decision epoch. A stationary deterministic Markovian policy $\pi=(d,d,\ldots)$ for an infinite-horizon MDP is a sequence of decision rules, where the deterministic decision rule does not change with the decision epochs. The value function $V^\pi :S \ra \R$ associated with a policy $\pi$ is the expected total discounted reward obtained by following the policy $\pi$ and is defined as
\begin{equation}
\label{eqn:vpi}
 V^{\pi}(s) =  \E{\left[\sum\limits_{t=0}^{\infty} \gamma^t R(s_t,d(s_t)) | s_0 = s\right]},
\end{equation}
for all $s \in S$. Here, $0 \leq \gamma < 1$ is the discount factor and it measures the current value of a unit reward that is received one epoch in the future.
The value function is the performance criterion to be optimized in the sequential decision-making problem modeled as MDP.
Thus, the objective is to find a stationary policy $\pi^* = (d^*,d^*,\ldots)$ such that 
\begin{equation}
\label{eqn:pistar}
 V^{\pi^*}(s) = \max_{\pi \in \Pi^{SD}} \;  V^{\pi}(s), \qquad \forall s \in S
\end{equation}
where $\Pi^{SD}$ is the set of all stationary deterministic Markovian policies. An optimal stationary deterministic Markovian policy satisfying \eqref{eqn:pistar} is known to exist under the following assumptions:
\begin{assumption}\label{as:rbdd}
$|R(s,a)| \leq C < \infty$, $\forall a \in A \; \forall s \in S$.
\end{assumption}
\begin{assumption}\label{as:statpr}
 Stationary $P$ and $R$, i.e., the functions $P$ and $R$ do not vary over time.
\end{assumption}
Dynamic programming (DP) \cite{BertB} techniques iteratively solve \eqref{eqn:pistar} and provide an optimal policy and the optimal value function for the given MDP based on the above assumptions, when model information in terms of $P$, $R$ is known. Model-free reinforcement learning (RL) \cite{sutton} algorithms on the other hand obtain the optimal policy when Assumptions 1 and 2 hold, but model information is not available (and not estimated). In the non-stationary environment scenario, as we consider, Assumption \ref{as:statpr} is invalid. Clearly, classical RL algorithms cannot help in learning optimal policies when Assumption 2 does not hold true. 
In the next section, we formally describe the problem of non-stationary environments and develop an RL algorithm which can tackle non-stationary environments.

%% file: pf-ai2019.tex
In this section, we formulate the problem of learning optimal policies in MDP environments with model changes and 
introduce the notation that will be used in the rest of the paper. 
We define a family of MDPs $\{M_\theta\}$, where $\theta$ takes values from a finite index set $\Theta$. For each $\theta \in \Theta$, we define
$M_\theta = \langle S, A, P_\theta , R_\theta \rangle$, where $S$ and $A$ are the state and action spaces, while $P_\theta$ is the transition probability kernel and $R_\theta$ is the reward function as defined before. The agent observes a sequence of states $\{s_t\}_{t \geq 0}$, where $s_t \in S$. For each state, an action $a_t$ is chosen based on a policy. For each pair $(s_t,a_t)$, the next state $s_{t+1}$ is chosen according to the active environment model. \emph{Changepoints} refer to the decision epochs at which the environment model changes. We denote the changepoints using the set of times $\{T_i\}_{i \geq 1}$. 
Here $\{T_i\}_{i \geq 1}$ form an increasing sequence of random integers. Thus, for example, at time $T_1$, the environment model will change from say $M_{\theta_0}$ to $M_{\theta_1}$, at $T_2$ it will change from $M_{\theta_1}$ to say
$M_{\theta_2}$ and so on, where $\theta_0, \theta_1,\ldots \in \Theta$. With respect to these model changes, the non-stationary dynamics for $t \geq 0$ will be 
\begin{gather}
P(s_{t+1} = s'|s_t = s,a_t = a) =    
\begin{cases}
  P_{\theta_0}(s,a,s'),\, \text{for } t < T_1\\    
  P_{\theta_1}(s,a,s'),\, \text{for } T_1 \leq t < T_2\\ 
  \vdots\\
\end{cases}
\end{gather}
and the reward for $(s_t,a_t) = (s,a)$ will be
\begin{gather}
R(s,a) = 
\begin{cases}
  R_{\theta_0}(s,a),\, \text{for } t < T_1\\    
  R_{\theta_1}(s,a),\, \text{for } T_1 \leq t < T_2\\    
  \vdots\\
\end{cases}
\end{gather}
We define the randomized history-dependent decision rule at time $t$ as $u_t : H_t \ra \mathcal{P}(A)$, where $H_t$ is the set of all possible histories at time $t$ and $\mathcal{P}(A)$ is the set of all probability distributions on $A$. An element of $H_t$ is of the form $h_t = (s_0,a_0,s_1,a_1,\ldots,s_t)$. $u_t$ is history dependent since the distribution $u_t(h_t) \in \mathcal{P}(A)$ picked is dependent on the sequence of states and actions observed upto time $t$. 
Given this rule, the next action at current state $s_t$ is picked by sampling an action from $u_t(\cdot)$.
If the decision rule is dependent on the current state only, irrespective of the history upto time $t-1$,
then $s_t$ suffices in place of $h_t$ and the decision rule is Markovian.
Deterministic Markovian decision rule $d_t:S \ra A$ defined earlier in the previous section, is then equivalent to $u_t$ when $H_t$ is replaced with $S_t$ and 
$\mathcal{P}(A)$ is replaced with the class of degenerate probability distributions over $A$. 

Given the family of MDPs $\{M_\theta\}$, the \emph{objective} is to learn a policy $\pi = (u_1,u_2,\ldots)$ such that the long-run expected sum of discounted rewards, i.e., $\E{ \left[ \sum\limits_{t=0}^{\infty} \gamma^t R(s_t,u_t(H_t)) | H_0 = h_0 \right]}$ is maximized for all initial histories $h_0 \in H_0$. 
However, multiple issues arise when Assumption \ref{as:statpr} is not satisfied which are compounded by the lack of model information:
\begin{enumerate}
 \item With changes in environment parameters, a stationary Markovian deterministic policy may not be optimal with respect to the above objective. Thus, any algorithm operating in this scenario needs to search over the space of randomized history-dependent policies which is an intractable problem. 

 \item When model information is not available, only samples of state and reward from simulation are available. A major problem is to use the state and reward samples to design an approximately optimal policy for non-stationary environments. Moreover, it is not clear as to which policy we must follow during the learning phase.
\end{enumerate}
In the next section, we explore these issues and provide solutions to address them. We mainly provide solutions for the RL setting when model information is unavailable, suggesting what policies to be used while learning and how to detect changes. Hence, our method can also be utilized in the setting when model information is available. In this case, as we will reason out later, since the agent can compute the optimal policies individually for all MDP models/contexts, it needs to only learn when the context change occurs.  

%% file: method-ai2019.tex
In this section, we describe the RL method to deal with changes in environment models. The method adapts a change detection algorithm \cite{prabukj} 
to find changes in the pattern of state-reward tuples observed while learning a policy.

\subsection{Experience Tuples} 
The change in $P$ and $R$ ultimately manifests itself in a change in sample paths. Taking a cue from this, we consider change detection on state-reward sequences.
By tracking variations in the state-reward sequence, our method 
captures variations in the occurrences of states and rewards, unlike previous works \cite{taposh1,hadoux}. We call each sample of state and reward in this sequence as an \emph{experience tuple}. Formally, 
an experience tuple $e_t$ at epoch $t$ is the triplet consisting of 
the current state $s_t$, current immediate reward (or cost) obtained $r_t$ and the next state $s_{t+1}$. So, $e_t = \langle s_t,r_t,s_{t+1}\rangle$. The set of experience tuples $\{e_t : 1 \leq t \leq B\}$, where $B$ is a batch size, is input to the changepoint detection algorithm \cite{prabukj}. 

For the proposed solution, we assume some structure in how the context changes occur. These assumptions are broadly mentioned below:

\begin{assumption}\label{as:pattern}
The environment model changes at least once. 
Additionally, the pattern of model change(s) is also known and the number of such changes is finite.
\end{assumption}
\begin{assumption}\label{as:suff_samples}
The environment context changes are not too frequent, i.e., we get sufficient state-reward samples for every context before the environment switches to some other context.
\end{assumption}


\subsection{Change Detection using Experience Tuples}
\label{subsubsec:cd}
We adapt the changepoint detection algorithm proposed in \cite{prabukj} for data consisting of experience tuples.
\cite{prabukj} describes an online parametric Dirichlet changepoint (ODCP) algorithm for unconstrained multivariate data. 
The ODCP algorithm transforms any discrete or continuous data into compositional data and utilizes Dirichlet parameter likelihood 
testing to detect change points. Multiple changepoints are detected by performing a sequence of single changepoint detections.

ODCP uses an appropriate metric while detecting change points.
One can also adapt other change detection algorithms like E-Divisive Change Point detection (ECP) \cite{ecp}. However, ECP uses Euclidean distance based metric to detect change points, which may not be 
suitable for discrete and compositional data that do not follow Euclidean geometry. Thus, ODCP reliably estimates change points compared to ECP.

ODCP requires the multivariate data to be i.i.d samples from a distribution. However, we utilize it in the Markovian setting, where the data obtained
does not consist of independent samples. But the following justification helps in understanding why adapting ODCP for experience tuples might still 
be a good idea: 
Let us suppose we choose the actions according to a stationary Markovian randomized policy $\pi = \{u,u,\ldots, \}$, where $u(s) \in \mathcal{P}(A)$.
With an abuse of notation, we denote each decision rule as $\pi$, with $\pi(s) \in \mathcal{P}(A)$.  
Let $\psi (a | s)$ be the probability that action $a$ is selected in state $s$ according to the decision rule $\pi(s)$.
Let ${\phi}^{\pi}(\cdot)$ denote the steady state distribution under policy $\pi$.
If we assume that the Markov reward process $\{P^\pi,R^\pi\}$ is fast mixing \cite{mixperes}, then we get experience tuples based on the steady-state distribution $\phi^\pi$.
Under this condition, the tuple $(s,r,s')$ namely the current state, reward and the next state will be distributed as follows:
$$(s,r,s') \sim {\phi}^{\pi}(s)\psi(\cdot | s)P(s,\psi (\cdot | s), s') .$$ 
The tuples $(s_t,r_t,s_{t+1})$ are identically distributed under  data and utilize the ODCP algorithm. 

We now provide some insight into the number of samples required for detecting changes in environment parameters using experience tuples. This is related to Assumption \ref{as:suff_samples}. The number of samples required is dictated by the size of the state and action space of the MDP. Let $m = |S|$, the size of the state space and $n = |A|$, the size of action space. Additionally, suppose $R^\pi(s,a) \in \mathcal{R}, \, \forall (s,a) \in S \times A$, where $\mathcal{R}$ is a finite set. The number of possible state-reward-state tuples will be $m\times m\times |\mathcal{R}|$. For efficient detection and reduced \emph{false alarm} probability, we need to get enough number of samples such that the state occupation probabilities are close to the actual steady-state probabilities.

The ODCP algorithm~\cite{prabukj} computes candidate changepoints by randomly permuting the given data samples. These candidate changepoints are ranked based on their statistical significance (see \cite{prabukj} for more details). The number of permutations to be tested with a candidate changepoint is prefixed and is chosen based on the number of data samples. Thus, more the number of samples, higher is the number of permutations tested. Hence the number of permutations to be fixed is based on the number of experience tuples the agent obtains for every context.

\subsection{Sampling Mechanism for Collecting Experience Tuples}\label{subsec:exp-policy}
In Section \ref{sec:pf}, we identified issues which arise when Assumption \ref{as:statpr} is violated. To address the issue of sampling experience tuples, we design a policy that the agent can follow to collect the experience tuples.
The RL method must detect a change in environment model (if it occurs) when the agent is in the process of learning a policy for controlling the MDP model. Thus, the \emph{behaviour policy} which the agent utilizes to explore the MDP model (like in QL~\cite{ql}, SARSA~\cite{sutton}) during learning, should also help the agent to get information about context (i.e., environment) changes. With this idea, we describe three mechanisms for exploration through behaviour policy:
\begin{enumerate}
 \item $\epsilon$-policy: Suppose the model information pertaining to all environment contexts is known, i.e., the agent knows $P_\theta$, $R_\theta$, $\forall \theta \in \Theta$. Hence, the optimal policies corresponding to all contexts is also known to the agent. However, in order to detect changes, there is a need for the agent to follow a policy which is approximately optimal. The reason to adopt an approximately optimal policy will be clear with the following example. Suppose that context changes from MDP $M_0$ to $M_1$ such that the probability and reward functions are same under the optimal policy of $M_0$, i.e., $P_0(s,\pi_0^*(s),s') = P_1(s,\pi_0^*(s),s')$ and $R_0(s,\pi_0^*(s),s') = R_1(s,\pi_0^*(s),s')$, but $\pi^*_1 \ne \pi^*_0$ and $\pi^*_1$ is optimal for $M_1$. Then by following the optimal policy $\pi^*_0$ for $M_0$, the agent will not be able to detect changes in the environment. This is because, the distribution of state-reward samples does not change under this policy $\pi^*_0$. So there is reason to explore other actions even though the optimal policy of both contexts is known. So, a sampling mechanism needs to explore actions other than the optimal action in order to detect changes. However, such exploration should be appropriately controlled, because there is the risk of following non-optimal actions quite often while controlling the MDP system.
As part of our solution we prescribe that experience tuples be collected using the following specific randomized policy when model information is known:
at each state $s$, the agent should follow optimal action prescribed by $\pi^*_i$ with probability $(1-\epsilon)$ and a random action with probability $\epsilon$, where $\epsilon > 0$. Thus, the policy used is $\pi = (u,u,\ldots)$, where $u:S \ra \mathcal{P}(A)$, $q_u(\pi^*_i(s)) = 1-\epsilon$ and 
$q_u(a) = \frac{\epsilon}{|A|-1}$, $a \in A\setminus \{\pi^*_i(s)\}$. We call this an $\epsilon$-policy. Here, based on Assumption \ref{as:pattern}, the agent knows the active context $i$.

\item Model-free RL policies:  The model information of contexts is not known. In this case, the agent can collect experience tuples while simultaneously following a model-free learning algorithm to learn an approximately optimal policy. We propose the use of Q-learning (QL) \cite{ql}, a model-free iterative RL algorithm to obtain 
the experience tuples. QL estimates the optimal Q-values for all state-action pairs of an MDP. The Q-value of a state-action pair w.r.t policy $\pi$ is defined as the expected discounted return starting from state $s$, taking action $a$ and following policy $\pi$ thereafter. The QL iteration \cite{ql} requires that all state-action pairs be 
explored for an infinite number of times, so that the optimal Q-value of each pair can be accurately estimated, based on the reward obtained at each step of the algorithm. 
To ensure this, an exploration strategy is used. As part of our solution, we prescribe that experience tuples be collected using either of the following strategies: 
\begin{itemize}
 \item $\epsilon$-greedy: In state $s$, with probability $(1-\epsilon)$, the action maximizing the Q-value estimate of state $s$ at iteration $k$, i.e., $\argmax_b Q_k(s,b)$ is selected, while with probability $\epsilon$, a random action in $A$ is selected. 
 \item UCB \cite{exp}: In state $s$, an action $a$ is selected as follows:
$$ a = \argmax\limits_{b}\left( Q_k(s, b)+  C \sqrt {\frac{\log N_k(s)}{N_k(s, b)}} \right),$$
where $Q_k(s,\cdot)$ is the estimate of the Q-value at iteration $k$, $N_k(s)$ tracks the number of times state $s$ is visited until $k$ and $N_k(s,b)$ is the number of times action $b$ has been picked when state $s$ is visited until $k$. Also, $C$ is a positive constant.
\end{itemize}
\end{enumerate}
Using the samples collected, the RL agent can detect changes, which can be carried out in an online fashion. In the following subsection, we describe how $\epsilon$-policy combined with ODCP can efficiently control an MDP system when model information is known.

\subsection{Leveraging Knowledge of Context Information}\label{subsec:ci}
When model information is known, the agent can compute a policy which will be optimal with respect to the objective defined in Section \ref{sec:pf}. However, as noted earlier, when Assumption \ref{as:statpr} is violated, a stationary Markovian policy need not be optimal when model parameters $P$ and $R$ change. Computing a non-stationary or non-Markovian policy is computationally infeasible, since this will involve search over an infinite set of policies. Moreover, the agent cannot employ standard dynamic programming techniques like value iteration and policy iteration \cite{puterman}  to compute the optimal policy, since these iterative methods have been designed based on the fact that optimal policies for stationary MDPs are stationary and Markovian. 

Our method is geared towards an alternate possibility. The autonomous agent can detect changepoints from the data comprising of experience tuples using the $\epsilon$-policy as described above. It can further use this information to compute the optimal policies for non-stationary MDP environments. The exact form of this non-stationary policy can also be described. The agent (based on Assumption \ref{as:pattern}), knows exactly which MDP context is active at the starting decision epoch. It begins to collect experience tuples using $\epsilon$-policy and simultaneously analyzes these samples for changes using ODCP algorithm. When the agent can sense that the context has changed, it switches to the optimal policy of the MDP context which is next in the known pattern of changes and which it presumes to be the current active context (based on the changepoint computed). After the policy switch, the agent continues to analyze the samples for changes. Then similar technique of switching continues for other contexts in the pattern as well. It is clear that this switching of policies gives rise to a non-stationary policy when all the individual policies are ordered together. Our method does not search over the large space of non-stationary, non-Markovian policies and instead gives a piecewise stationary policy that the agent can easily compute.

\subsection{Context Q-learning}\label{subsec:cql}
For the scenario where model information of all contexts is known, the method just described is useful to obtain a piecewise stationary policy. However, in the case when context information is not known, we need to design a method for finding the optimal policy. Such a method should be sensitive to environment changes as well.   
We design \emph{Context Q-learning (Context QL)}, which is a method that can handle the learning task when model/context information is not known. The concept of Context QL is in one respect similar to RLCD \cite{rlcd}. Both methods instantiate new models whenever a change is detected. However, unlike RLCD which utilizes a heuristic quantity for tracking and declaring changepoints, our method works in tandem with a changepoint detection algorithm, ODCP, to get information about the changes in contexts. 
Furthermore, Context QL updates Q values of the relevant model whenever a change is detected and does not attempt to estimate the transition and reward functions for the new model. Additionally, if the method finds that samples  are obtained from a previously observed model, it updates the Q values corresponding to that model. Thus, in this manner, the information which was learnt and stored earlier (in the form of Q values) is not lost. 
The Context QL pseudocode is given in Algorithm \ref{alg:cql}.

The Context QL algorithm takes as input the pattern of changes in the environment models $M_1,\ldots,M_k$, so that the Q values of the right model are updated. It is to be noted that only the change pattern is known and not the context information.
However, when these model changes occur, the changepoints $T_1,\ldots,T_n$ are not known to Context QL.
For example, suppose the agent knows that model changes from say $M_0$ to $M_1$ and then to $M_2$. This implies that the agent 
knows that dynamics change from $M_0$ to $M_1$, but it does not know $P_0$, $R_0$, $P_1$, $R_1$ etc.
Based on the pattern information, Context QL updates Q values pertaining to model $M_0$ initially. Later when first change is detected, it updates Q values of $M_1$, followed by updates to Q values of $M_2$ when another change is detected.
The algorithm initializes a context counter $c$, which keeps track of the current active context, according to the changes detected. It maintains Q values for all known contexts $1,\ldots,k$ and initializes the values to zero. The learning begins by obtaining experience tuples $e_t$ according to the dynamics and reward function of context $M_{\theta_1}$. The state and reward obtained are stored as experience tuples, since model/context information is not known. The samples can be analyzed for context changes in batch mode or online mode, which is denoted as a function call to ODCP in the algorithm pseudocode. If ODCP detects a change, then the counter $c$ is incremented, signalling that the agent believes that context has changed. The lines 6-16 represent this learning phase when context $M_{\theta_1}$ is active.
Similar learning takes place for other contexts as well.
\begin{algorithm}\caption{Context QL}
\begin{algorithmic}[1]
\STATE {\textbf{Input:}} Model change pattern, $M_{\theta_1} \rightarrow M_{\theta_2}$, $M_{\theta_2} \rightarrow M_{\theta_3}$, $\ldots, M_{\theta_{n-1}} \rightarrow M_{\theta_n}$, where $M_{\theta_i} \in \{M_1,M_2,\ldots,M_k\}$
and $\theta_i \in \Theta = \{1,2,\ldots,k\}$
\STATE Fix learning rate $\alpha$
\STATE {\textbf{Initialization:}} Context number, $c = 1$, Q values $Q(m,s,a) = 0$, $\forall m \in {1,\ldots,k}$, $\forall (s,a) \in S \times A$
\STATE {\textbf{Initialization:}} Initial state $s_1 = s$, $\tau^* = 1$, $T_0 = 0$
\FOR{$L=1$ to $n$}
\FOR{$i= T_{L-1}$ to $T_{L}$}
    \STATE Follow action $a_i$ according to $\epsilon$-greedy or UCB exploration
    \STATE Obtain next state $s_{i+1}$ according to $M_{\theta_{L}}$ dynamics
    \STATE Get reward $r_i$ according to $M_{\theta_L}$ reward function
    \STATE Update Q value $Q(j_c,s_i,a_i)$ as follows:
        \begin{equation*} Q(\theta_c,s_i,a_i) := (1-\alpha) Q(\theta_c,s_i,a_i) + \alpha (r_i + \gamma \max_b Q(\theta_c,s_{i+1},b)) \end{equation*}
    \STATE $e_i = \langle s_i,r_i, s_{i+1}\rangle$
    \STATE $\tau= $ ODCP$(\{e_t : \tau^* \leq t \leq i\})$
    \IF{$\tau$ is not Null}
        \STATE Increment $c$
        \STATE $\tau^* = \tau $ 
    \ENDIF
    \STATE $i = i + 1$
\ENDFOR
\ENDFOR
\end{algorithmic}
\label{alg:cql}
\end{algorithm}
In line 1, the input uses the notation $M_{\theta_i}$ to indicate that the system can initially evolve according to any of the $k$ models and context $M_0$ need not be the initial active context. Thus, $1 \leq \theta_i \leq k$.
The information $T_1, T_2, \ldots, T_n$ is known only to the environment. In Algorithm \ref{alg:cql} it is shown as though the RL agent is aware of it. But this is not so, it has been included only to indicate that at decision epochs independent of the agent, the context changes.

\begin{remark}\label{rem:avg}
In Section \ref{sec:pf} and here, we have mentioned that given a family of contexts, $\{M_\theta\}$, the \emph{objective} is to 
learn a policy $\pi = (u_1,u_2,\ldots)$ such that the expected sum of discounted rewards accumulated over the infinite horizon
i.e., $\E{ \left[ \sum\limits_{t=0}^{\infty} \gamma^t R(s_t,u_t(H_t)) | H_0 = h_0 \right]}$ is maximized for all initial histories $h_0 \in H_0$.
In this Section,  we described our method, Context QL, which finds optimal policies for each environment 
context that maximizes (or minimizes) the long-run discounted reward (or cost) criterion.
It should be noted that, if, instead we have the average reward criterion for the policy wherein the objective is to maximize  
$g^{\pi}(s) = \lim\limits_{N \ra \infty} \frac{1}{N} \E \left[ \sum\limits_{t=1}^{N} R(s_t,d(s_t)) | s_0 = s\right]$, $\forall s \in S,$ over all policies $\pi \in \Pi^{SD}$, then our method can still be utilized. 
This is possible if we adapt the relative Q-value iteration algorithm \cite{rvqi} to non-stationary settings.  
The Context QL method can be easily extended to the average reward per step setting in this manner.
\end{remark}

\subsection{Performance Evaluation and Policies}
To evaluate our method, we need a performance metric. In the classical case of stationary MDPs, the RL algorithms learn a policy (maybe deterministic or randomized) and the algorithms are evaluated based on the rewards these policies yield when they are used to control the MDP model.
So, for the non-stationary algorithms proposed, in a similar fashion, we compute the rewards garnered by the algorithms. Sections \ref{subsec:rmdp}-\ref{subsec:tr} show the performance of Context QL w.r.t the cumulative reward gathered when the model information is not available. 
This is in contrast with prior works like UCRL2 \cite{ucrl2}, variation-aware UCRL \cite{gajane2019paper1}, where \emph{regret} is the performance metric (see \cite{ucrl2,gajane2019paper1} for a formal definition). \emph{Regret} is a measure of the rewards which are missed by the policies yielded by these algorithms when compared to the optimal policy. In order to compute this performance metric, we ought to know the optimal policy and its performance. Clearly, the idea of computing \emph{regret} is not applicable to model-free RL, where model information is not known. Thus, in our view, measuring \emph{regret} is not apt for the problem we consider. However, when model information is known, regret can be computed. We show some results for this case in Section \ref{subsec:rmdp}.

With non-stationary environments, Assumption \ref{as:statpr} does not hold true. Hence, there is no reason to believe that a stationary policy is optimal when this assumption is not satisfied in a system. There is a need to search the large space of non-stationary policies for ensuring optimal behaviour of the MDP system model in dynamic environments. An exhaustive search over this large space is computationally expensive and we require an algorithm which can find an approximately optimal policy over a smaller, restricted set. The Context QL algorithm discussed in Section \ref{subsec:cql} does the same. The smaller, restricted set of non-stationary policies we are interested in is the set of piecewise-stationary policies, as described in Section \ref{subsec:ci}.
We evaluate the experience tuples for changes in the underlying distribution and find the changepoints. Before the change occurs, the RL agent follows a stationary optimal policy learnt for the respective context. After the change, the RL agent switches and learns another stationary policy. Also, if the environment was previously observed, then the RL agent updates the policy learnt for the same environment, by utilizing the additional samples obtained.

\subsection{Scalability of Context QL}
Context QL (Algorithm \ref{alg:cql}) stores Q values for every context. From the learnt Q values, a policy is derived~\cite{ql} and followed. However, this memory requirement is independent of the number of changes in the environment. For e.g., if $\theta_i \in \Theta = \{1,2\}$ and the model change pattern is $M_{\theta_1} \ra M_{\theta_2} \ra M_{\theta_1} \ra M_{\theta_2} \ra M_{\theta_1} \ra M_{\theta_2}$, the number of Q tables stored will be two, one for each of the models $M_{\theta_1}$ and $M_{\theta_2}$. In this example the number of changes is $5$, which does not affect learning of either of the two models. However, the learning accuracy and the policy learnt is directly dependent on how many samples are obtained for each of the contexts/models.

%% file: expres-ai2019.tex
In this section, we evaluate our method for accuracy in the changepoints detected and the reward 
accrued.
The experience tuples are collected from randomly-generated MDPs (in Section \ref{subsec:rmdp}), from a sensor application (see Section \ref{subsec:sen})
and a traffic application (see Section \ref{subsec:tr}).
All numerical experiments are carried out using R statistical package and Python programming language.

We also compare the accuracy of changepoints detected by ODCP \cite{prabukj} and E-Divisive (ECP) \cite{ecp} for 
the data consisting of experience tuples. ECP is a changepoint detection algorithm. It is seen to perform well on many synthetic multivariate datasets. Numerical simulations described in~\cite{ecp} show that it can detect changes in mean, variance and covariance of dataset. Additionally, it is seen to perform well on many real datasets \cite{ecp}.

\subsection{Random MDP}\label{subsec:rmdp}
 We test our method on different Random MDP models generated using MDPtoolbox \footnote[1]{https://cran.r-project.org/web/packages/MDPtoolbox}. First, the methods are tested for single changepoint detection followed by multiple changepoints detection.
 The results below are grouped according to this. In all experiments, the discount factor $\gamma$ is set to $0.9$.
\subsubsection{Single Changepoint Detection} 
We consider model change from $M_0$ to $M_1$ and collect $2000$ samples (i.e., the three-dimensional experience tuples). 
The actual model change occurs at $T_1 = 1000$. 
  The first $1000$ state-reward samples are from $M_0$ and the rest $1000$ samples are from $M_1$.
  Let $\tau^*$ be the changepoint detected by the various methods.
Table \ref{table:perf-model-based} summarizes the changepoints detected when we assume that model information is known and the agent needs to determine when the environment switches from $M_0$ to $M_1$. Note that, as remarked in Section \ref{sec:method},  the best possible alternative to using a non-stationary policy in this scenario is to switch to an appropriate stationary policy.
Since model information is known, optimal policies for both environments are also known and the agent can switch between these optimal policies if it can reliably determine the changepoint. In order to detect changes, the samples are collected using a randomized policy with $\epsilon = 0.1$, as described in Section \ref{subsec:exp-policy}. 

Table \ref{table:perf} summarizes the changepoints  detected when we assume that model information is not known. Thus, in addition to determining the changepoints, the agent also needs to estimate the optimal policies for the environments. In order to detect changes, the samples are collected using $\epsilon$-greedy and UCB policies.
 \begin{table}[h]
    \begin{center}
    
        \begin{tabular}{|c|c|c|c|}
        \cline{1-4}
        Change Detection Algorithm      &   Mean of $\tau^*$        & SD of $\tau^*$             &    Median of $\tau^*$ \\        
        \cline{1-4}
        ODCP with $\epsilon$-policy     & $999.3$                   & $25.88659$                 &     $1001.5$   \\
        \cline{1-4}
        ECP with $\epsilon$-policy      & $1008.65$                 & $32.21029$                 &     $1002$  \\
        \cline{1-4}
        \end{tabular}     
         \caption{\small Performance comparison of ODCP and ECP in changepoints detected when model information is known.}
        \label{table:perf-model-based}
       
    \end{center}
 \end{table} 
 \begin{table}[h]
    \begin{center}    
   
        \begin{tabular}{|c|c|c|c|}
        \cline{1-4}
        Method                                      &   Mean of $\tau^*$            & SD of $\tau^*$             &    Median of $\tau^*$ \\       
        \cline{1-4}        
        ODCP, QL with UCB                           & $1001.11$                        & $31.86$                       & $997$  \\
        \cline{1-4}
        ECP, QL with UCB                            & $1006.22$                        & $36.65$                      & $1001$  \\
        \cline{1-4}
        ODCP, QL with $\epsilon$-greedy             & $986.9444$                    & $34.769$                   & $999$   \\
        \cline{1-4}
        ECP, QL with $\epsilon$-greedy              & $1027.833$                    & $59.69$                   & $1002$  \\
        \cline{1-4}
        
        \end{tabular}     
       
         \caption{\small Performance comparison of ODCP and ECP in changepoints detected when model information is not known.}
        \label{table:perf}
    \end{center}
 \end{table}

 In Tables \ref{table:perf-model-based} and \ref{table:perf}, the mean, median and standard deviation of $\tau^*$ (over $20$ sample trajectories) is presented. It can be observed from Table \ref{table:perf-model-based} and Table \ref{table:perf} that ODCP and ECP detect change in model using just tuples (changes manifest in tuples), promising that experience tuples can be utilized to detect changes in environment contexts. 
As can be inferred from these results, the average and median of $\tau^*$ found by ECP and ODCP are very close, but mean ODCP performance is better compared to ECP. However, it is observed that when experience tuples are analyzed for changepoints, ECP has a higher standard deviation in the changepoints detected when compared to ODCP. In Table \ref{table:perf}, ODCP with QL and ECP with QL utilize QL to learn policies, but these do not maintain separate Q-values for the environments.
This table also shows results with two exploration strategies. In Table \ref{table:perf}, UCB exploration is much better than $\epsilon$-greedy in both ODCP and ECP algorithms. This is because of the constant value of $\epsilon$. Even after optimal actions are learnt for the various states, exploratory actions would continue to be chosen. This is not so with UCB-based algorithms. Also, ODCP shows better results on the whole as compared to ECP.

For the case when model information is not known, we also implemented RLCD \cite{rlcd} and its extension \cite{hadoux} to detect changes in the environment. These algorithms rely on tracking the quality of the model learnt. If the quality crosses some threshold, a new model is instantiated and estimated. 
The choice of this threshold is crucial. In our simulations, we observed that RLCD instantiates more than four environment models, even though there is a single change in the context. The same results were observed with \cite{hadoux}. Due to this reason, the results pertaining to RLCD and \cite{hadoux} were not included in Table \ref{table:perf}.

In Tables \ref{table:precision-recall-w100}, \ref{table:precision-recall-w50} we analyze the precision and recall~\cite{precrecall} metric for algorithms proposed in~\cite{taposh1,tnnls} and compare it with the performance of Context QL with respect to this metric.
For this, the model information is known and all policies are computed based on the context information. The two-threshold switching strategy proposed in~\cite{taposh1} utilizes the known transition probabilities of all contexts to compute the Shiryaev-Roberts statistic (SR)~\cite{shiryaev} for every decision epoch. This value is initialized to $0$ and is updated as follows:
\begin{equation*}
    SR_{t+1} = (1 + SR_{t}) \frac{P_{\theta_1}(s,a,s')}{P_{\theta_0}(s,a,s')}, \quad \forall t >0.
\end{equation*}
If SR is lower than a prefixed threshold $B$, then optimal policy $\pi^*_{\theta_0}$ corresponding to the MDP model $M_{\theta_0}$ is followed. However, if SR is greater than $B$, but lower than another prefixed threshold $A > B$, then a policy $\pi_{KL}$ as designed below is followed:
\begin{equation*}
    \pi_{KL}(s) = \argmax\limits_{a \in A} \sum\limits_{s'} P_{\theta_1}(s,a,s') *\log\left(\frac{P_{\theta_1}(s,a,s')}{P_{\theta_0}(s,a,s')}\right), \quad \forall s \in S.
\end{equation*}
The above policy $\pi_{KL}$ is designed for better detection of changes, though it may not be an optimal policy for model $M_{\theta_0}$. Once SR crosses the threshold $A$, the optimal policy 
$\pi^*_{\theta_1}$ is followed. The key feature of the two-threshold strategy is to prefix appropriate values of $A$ and $B$.~\cite{taposh1} does not provide any method for identifying these values and hence in the experiments, we have shown results of precision and recall with multiple sets of these values.
The P-CDM and NP-CDM algorithms proposed in~\cite{tnnls} detect changes in the transition probability of discrete-time Markov chains (DTMCs). We adapt these to find concept drift in MDPs when context information is known. MDP context information being known implies that the optimal policies of each MDP model can be computed. When an optimal policy is followed, we obtain a Markov chain for the corresponding MDP model. For this chain, we further adapt the algorithms P-CDM and NP-CDM to find the changepoints. P-CDM and NP-CDM algorithms compute the stationary distribution $\xi_j$ for context $M_{\theta_j}, \; \forall j \in \{1,2,\ldots,k\}$. Both algorithms consider the non-overlapping sequences of states which evolve based on the known transition probabilities. A moving window $w_i$ of state samples at epoch $i$ comprises of the states $\{s_{(w_{i-1}+1)},\ldots,s_{w_i}\}$. P-CDM and NP-CDM differ only in the manner in which this sequence is picked. For the sequence picked, P-CDM and NP-CDM compute the log-likelihood ratio $l_i = \log\left(\frac{\mathbb{P}_{\theta_1}(w_i)}{\mathbb{P}_{\theta_0}(w_i)}\right)$, where $$\mathbb{P}_{\theta_j}(w_i) = \xi_j(s_{(w_{i-1}+1)}) \prod_{k = w_{i-1}+1}^{w_i - 1} P_{\theta_j}(s,\pi^*_{\theta_j},s').$$
These algorithms track the sign of $l_i$ and compute $m_i = \max(0,m_{i-1}+\sign(l_i))$. If $m_i \geq K$ where $K$ is a pre-fixed threshold value, then a change is declared. In the numerical simulations, we vary $K$ and note the performance of P-CDM and NP-CDM w.r.t the precision and recall metric. 
For the numerical experiments in Tables \ref{table:precision-recall-w100}, \ref{table:precision-recall-w50} a datapoint being labelled as a changepoint is said to be correct if it lies within a pre-defined window size $W=100$, $W = 50$ respectively, of the actual, true changepoint. Based on this criterion, for each of the algorithms, the results in Tables \ref{table:precision-recall-w100}, \ref{table:precision-recall-w50} are obtained by averaging over $20$ Monte Carlo simulations, i.e., we sample $20$ sample paths randomly for performing Monte-Carlo averaging.
\begin{table}[h!]
\begin{center}%
\begin{tabular}{|c|c|c|c|c|c|c|c|c|c|c|c|c|}
\cline{1-12}
\multicolumn{3}{|c|}{P-CDM} & \multicolumn{3}{c|}{NP-CDM} & \multicolumn{4}{c|}{Two-threshold Switching Strategy} & \multicolumn{2}{c|}{Context QL}\\\cline{1-12}
K & Prec. & Recall & K  & Prec. & Recall & A       &    B   & Prec. & Recall & Prec. & Recall \\\cline{1-12}
5 &  1     &  0.7   & 5   & 0.1      &  1      & 100    & 50    & 0.05   & 1    &    &  \\\cline{1-10}
10  &  1     & 0.05  & 10 & 0.6   & 0.642  & 1000  & 500  & 0.706     & 0.8    &  1  &  0.75     \\\cline{1-10}
15  &  0  &  0  & 15 & 0.8125  & 0.7647  & 10000  & 5000  & 0.75     & 0.333    &       &        \\\cline{1-10}
20  &   0    &  0    & 20 & 0.8235  & 0.8235  & 100000  & 50000  & 1     & 0.2    &       &        \\\cline{1-10}
\hline  
\end{tabular}
\caption{\small The precision and recall metric for the algorithms P-CDM, NP-CDM proposed in \cite{tnnls}, Two-threshold Switching Strategy proposed in \cite{taposh1} and Context QL with window size $W = 100$. Variation of the precision and recall metric is seen based on the variation in the threshold values $K$ for P-CDM and NP-CDM and $A$ and $B$ for \cite{taposh1}.  }
\label{table:precision-recall-w100}
\end{center}
\end{table}

From Table \ref{table:precision-recall-w100}, we observe that P-CDM shows very poor performance, when compared to the other algorithms. With a smaller window size, this performance is only expected to degrade and hence results of P-CDM are not included in Table \ref{table:precision-recall-w50}. It is seen that with increase in the value of $K$, the performance of NP-CDM improves. However, similar to~\cite{taposh1}, where values of $A$ , $B$ are fixed after trial and error, the values of $K$ are also pre-fixed in a similar manner - this implies there is no single value of $K$ which gives good results for all sizes of state-action space. This can be inferred from Table \ref{table:s-a-prec-recall}, where we used $K = 20$ for NP-CDM. Though this value gives good precision and recall values with $|S| = |A| = 5$, the same value, when used for larger state-action space MDPs degrades the performance of NP-CDM. This can be observed in Table \ref{table:s-a-prec-recall}. With a smaller window, the precision and recall values are expected to decrease, which is observed in Table \ref{table:precision-recall-w50}. Note that, since Context QL has no threshold values to be tuned, we have a single row for Context QL in both Tables \ref{table:precision-recall-w100} and \ref{table:precision-recall-w50}. The identical row for Context QL in both Tables \ref{table:precision-recall-w100} and \ref{table:precision-recall-w50} indicates that the changepoints detected by Context QL lies within the smaller window size of $50$.
\begin{table}[h!]
\begin{center}%
\begin{tabular}{|c|c|c|c|c|c|c|c|c|c|}
\cline{1-9}
 \multicolumn{3}{|c|}{NP-CDM} & \multicolumn{4}{c|}{Two-threshold Switching Strategy} & \multicolumn{2}{c|}{Context QL}\\\cline{1-9}
K  & Prec. & Recall & A       &    B   & Prec. & Recall & Prec. & Recall \\\cline{1-9}
5   & 0.1      &  1      & 100    & 50    & 0.05   & 1    &    &  \\\cline{1-7}
10 & 0.6   & 0.642  & 1000  & 500  & 0.533     & 0.6154    &   1  &  0.75     \\\cline{1-7}
15 & 0.8125  & 0.7647  & 10000  & 5000   & 0.6     & 0.1667  &       &        \\\cline{1-7}
20 & 0.4  & 0.1176  & 100000  & 50000  & 0     & 0   &       &        \\\cline{1-7}
\hline  
\end{tabular}
\caption{\small The precision and recall metric for NP-CDM, Two-threshold Switching Strategy and Context QL with window size $W = 50$. Variation of the precision and recall metric is seen based on the variation in the threshold values $K$ for NP-CDM and $A$ and $B$ for the two threshold switching strategy.  }
\label{table:precision-recall-w50}
\end{center}
\end{table}

Table \ref{table:s-a-prec-recall} shows the variation in precision and recall metric of Context QL, two-threshold switching strategy and NP-CDM with respect to the variation in the size of state-action space of MDPs. We have set a window size of $W = 100$ for these simulations. Based on Table \ref{table:precision-recall-w100}, we set $K = 20$ for NP-CDM and $A = 1000$, $B = 500$ for two-threshold switching strategy. Though we have inferred that no single value of $K$, $A$, $B$ is apt for these algorithms, we have used a single value for the experiments in Table \ref{table:s-a-prec-recall}, since our focus is to show the changes in precision and recall metric with changes in state-action space.
\begin{table}[h!]
\begin{center}%
\begin{tabular}{|c|c|c|c|c|c|c|c|}
\cline{1-8}
\multicolumn{1}{|c|}{$|S|$} & \multicolumn{1}{c|}{$|A|$}  & \multicolumn{2}{c|}{Context QL} & \multicolumn{2}{c|}{Two-threshold Switching Strategy} & \multicolumn{2}{c|}{NP-CDM} \\\cline{3-8}
  &   & Prec. & Recall & Prec. & Recall & Prec. & Recall \\\hline

5  & 5  &  1    &  0.75     & 0.706 & 0.8  & 0.8235 & 0.8235 \\\cline{1-8}
8  & 8  &  1     &   1     & 0 & 0  & 0.8 & 0.706 \\\cline{1-8}
12 & 12 & 1      &   1     & 0 & 0  & 0.723 & 0.632 \\\cline{1-8}
\hline  
\end{tabular}
\caption{\small The precision and recall metric for the algorithms Context QL, Two-Threshold Switching Strategy~\cite{taposh1} and NP-CDM proposed in \cite{tnnls} for various sizes of state and action spaces.}
\label{table:s-a-prec-recall}
\end{center}
\end{table}
In Table \ref{table:s-a-prec-recall}, we see that performance of two-threshold switching strategy and NP-CDM degrades with increase in size of state-action space. This degradation is gradual in NP-CDM, however with two-threshold strategy, there is a drastic decrease in the performance. This is due to the fact that the pre-fixed threshold values which are apt for $|S| = |A| = 5$, are not apt for $|S|=|A|=8$ and $|S|=|A|=12$. However, for other sizes of state-action space, we do not know how to pre-determine and fix  appropriate threshold values.

Next, in Table \ref{table:reward-sum}, we provide results for the rewards collected by QL,  Context QL, UCRL2~\cite{ucrl2}, RLCD~\cite{rlcd} and RUQL~\cite{ruql} when the model information is not known. These results show the performance of these algorithms when there is a change in the environment from $M_0$ to $M_1$ at $T_1 = 1000$ with total number of experience tuples collected being $2000$. As a reference, we also provide the reward collected by ODCP-based and ECP-based methods when model information (and hence optimal policies) is known for both environments. The ODCP-based and ECP-based methods use $\epsilon$-policy for detecting changes and control MDP using optimal policy of $M_0$ followed by that of $M_1$ after a change is detected. 
Suppose the changepoint detected by ODCP is $\tau^*$. Let $\pi_0^*$ and $\pi_1^*$ be the optimal policies of $M_0$ and $M_1$ respectively. The ODCP and ECP methods with $\epsilon$-policy, control the MDP using policy $\pi_0^*$ from epoch $1$ to epoch $\tau^*-1$ and from epoch $\tau^*$ these pick actions according to $\pi_1^*$. 

QL does not detect changepoints and instead updates the same set of Q values, unlike Context QL (as described in Algorithm \ref{alg:cql}). RUQL is similar to QL, but its exploration strategy and step-size schedule differ from QL. UCRL2 estimates the upper confidence bounds on the transition probability and reward functions as described in \cite{ucrl2}. Also, we restart UCRL2 algorithm when the number of steps satisfies the criterion described in Theorem 6 of \cite{ucrl2}.
\begin{table}[h]
\begin{center}%
\begin{tabular}{|c|c|c|}
\hline
Method                       &   Mean $\pm$ SD              & Median        \\\hline
ODCP $\epsilon$-policy       &   $538.67 \pm 14.28$         & $538.0404$        \\\hline
ECP $\epsilon$-policy        &   $515.1421 \pm 23.26$       & $520.2686$    \\\hline
Context QL                   &   $365.6 \pm 64.14$          & $362.06$      \\\hline
QL                           &   $287.4 \pm 69$             & $284.4$          \\\hline
UCRL2                        &   $176.83\pm 11.25$          & $178.3511$        \\\hline
RUQL                         &   $103.41 \pm 26.6$          & $99$              \\\hline
RLCD                         &   $91.3341 \pm 20.77$          & $97.41$             \\
\hline  
\end{tabular}
\caption{\small Rewards collected by UCRL2, RLCD, RUQL, QL and Context QL. ODCP and ECP collected samples using $\epsilon$-policy.}
\label{table:reward-sum}
\end{center}
\end{table}

As observed from Table \ref{table:reward-sum}, RLCD and RUQL perform poorly when compared to QL, UCRL2 and Context QL in terms of the total reward obtained. In UCRL2 though the reward obtained is less when compared to QL and Context QL, the standard deviation is much lower when compared with QL and Context QL. However, unlike QL and Context QL, it has high space complexity, since it maintains estimates of probability and reward function along with various state-action counters required for the upper confidence bounds.

In the numerical experiments of Table~\ref{table:reward-sum}, we used $\epsilon$-greedy policy exploration for QL and Context QL. QL utilizes one set of Q values to learn optimal policies for contexts $M_0$ and $M_1$. It does not detect changes, but updates Q values using rewards from both the environments. This is where the issue arises. When the environment changes at $T_1$ from model $M_0$ to $M_1$, the Q values would have been updated using rewards from $M_0$. Once the environment starts providing samples from context $M_1$, the action choice of the agent is biased by the already updated Q values. This does not occur in the Context QL, since once a change is detected, Context QL starts updating another set of Q values for the new environment.

The performance of algorithms w.r.t the regret criterion is shown in Table \ref{table:regret1}, when the changepoints obtained are analyzed by ODCP and ECP.
The regret is computed for each sample trajectory comprising of $2000$ experience tuples, with a changepoint at $T_1 = 1000$.
The results in Table \ref{table:regret1} are computed over $20$ such sample trajectories.
\begin{table}[h]
\begin{center}

\begin{tabular}{|c|c|c|}
\hline
Method                                    &   Mean $\pm$ SD    & Median        \\\hline
Model-based, ODCP $\epsilon$-policy       &   $20 \pm 15$      & $16.6$        \\\hline
Model-based, ECP $\epsilon$-policy        &   $39 \pm 26$      & $52$    \\
\hline  
\end{tabular}
\caption{Regret of ECP and ODCP $\epsilon$-policy based method.}
\label{table:regret1}
\end{center}
\end{table}

\subsubsection{Multiple Changepoints Detection} 
We evaluate the accuracy of ODCP~\cite{prabukj} and ECP~\cite{ecp} on MDP for multiple changepoints. In the experiments,  the model alternates thrice between $M_0$ and $M_1$ starting with $M_0$. 
With $2000$ samples, changepoints are fixed at $T_1 = 500$, $T_2 = 1000$ and $T_3 = 1500$. 
Averaged over $20$ Monte Carlo simulations, mean of $\tau^*_1 = 520$, mean of $\tau^*_2 = 1059$ and mean of $\tau^*_3 = 1510$ for our method, while ECP identifies only the first changepoint with mean $855$. 
More importantly, ECP fails to detect the second and third changepoints. The reliability with which ODCP detects changepoints is better compared to ECP. 
In this setting of multiple changepoints, we also compared the performance of RLCD \cite{rlcd} with ODCP. The RLCD algorithm detects more than $3$ changepoints in this scenario.

In Table \ref{table:multiple-cps-reward-sum}, we show the total reward gained by the policies learnt by algorithms QL, UCRL2, RUQL and Context QL. For this, the samples were obtained from a Random MDP system. The MDP environment alternates between $M_0$ and $M_1$ twice, i.e., the sequence of context changes is $M_0 \rightarrow M_1  \rightarrow M_0  \rightarrow M_1$. QL and RUQL maintain a single set of Q values to learn the policy and UCRL2 is simulated with restarts as described in Theorem 6 of \cite{ucrl2}. Context QL maintains two different estimates of the Q values - one for each MDP environment and resumes updating the appropriate Q values when a change is detected. For the numerical experiments, we get $T = 4000$ $(s,s',r)$ samples, with $T_1 = 1000$, $T_2 = 2000$ and $T_3 = 3000$. Thus, at $T_1$, the MDP model changes from $M_0$ to $M_1$. At $T_2$, it again flips to $M_0$ and so on.

\begin{table}[h]
\begin{center}%
\begin{tabular}{|c|c|c|}
\hline
Method                       &   Mean $\pm$ SD              & Median        \\\hline
Context QL                   &   $945.94 \pm 146.54$          & $959.92$      \\\hline
QL                           &   $866.9 \pm 89$             & $882.2$          \\\hline
UCRL2                        &   $439.78\pm 27.31$          & $436.98$        \\\hline
RUQL                         &   $253.18 \pm 40$          & $255.54$              \\
\hline  
\end{tabular}
\caption{\small Rewards collected by UCRL2, RUQL, QL and Context QL. The environment model alternates between contexts $M_0$ and $M_1$.}
\label{table:multiple-cps-reward-sum}
\end{center}
\end{table}

\subsection{Energy Management in a Single Sensor Node with Finite Buffer}\label{subsec:sen}
\input{expres-sensor}

\subsection{Traffic Signal Control}\label{subsec:tr}
\input{expres-traffic}

\subsection{Summary of Results}\label{subsec:summary}
As observed in Tables \ref{table:perf-model-based}-\ref{table:traffic}, the results of our method are promising. Tables \ref{table:perf-model-based}, \ref{table:perf} show the detection delay of the model-based ODCP and ECP algorithms when adapted to the RL setting. Their performance is comparable for both settings - when model information is known and when it is not known. Both methods detect changes with least detection delay as can be judged from the mean, median and standard deviation values shown in these tables. The \emph{false positive}, \emph{false negative} errors and the \emph{true positive} values can be inferred from Tables \ref{table:precision-recall-w100}, \ref{table:precision-recall-w50} and \ref{table:s-a-prec-recall} which indicate the precision and recall metric for NP-CDM~\cite{tnnls}, two-threshold switching strategy~\cite{taposh1} and Context QL. 
It can be seen that Context QL is quite robust to the size of state-action space, has good performance and does not require tuning of any threshold values, unlike NP-CDM or the two-threshold switching strategy. Tuning these values requires model information and lot of trial and error. With real-life applications, one may not have model information. Tuning the values by trial and error may not be even feasible in real-life applications. Thus, Context QL presents a great advantage when the algorithm is to be used in real-world applications.

Tables \ref{table:reward-sum},  \ref{table:multiple-cps-reward-sum} show the average total reward obtained by QL, RUQL, UCRL2, RLCD and Context QL. 
The gap in total reward obtained by QL and Context QL is explicit in the multiple changepoints case where the contexts change in the fashion $M_0 \rightarrow M_1 \rightarrow M_0 \rightarrow M_1$. This is because, Context QL has separate Q value data structure for each environment. So, after the second and third changepoints, it faces the same environment that was observed earlier. The samples obtained after 2nd changepoint (i.e., after $\tau^*_2$) are used to update Q values of $M_0$ again. Similarly, after the 3rd changepoint (i.e., after $\tau^*_3$), the reward samples are used to update Q values of $M_1$ again. Hence even though the environment alternates between $M_0$ and $M_1$ twice, for each environment, we get almost equal number of reward samples.


In the experiments, we tested Context QL on two realistic applications - one in energy management in energy harvesting sensors and the other in traffic signal duration control. These are applications where the effect of changing environment parameters or system operating conditions is clearly visible. 
It is seen in Tables \ref{table:sensor} and \ref{table:traffic} that Context QL performs much better when compared to classical QL. 
Context QL captures the change in joint distribution of $(s,a,r)$ tuples.
When this distribution changes, the ODCP indicates that a changepoint is detected. This detection is based on permutation tests in ODCP and is hence flexible.
It should be noted that Assumptions \ref{as:pattern}, \ref{as:suff_samples} are important for our method. These assumptions are reasonable, because it is often observed that in applications like traffic signal control and sensor energy management, although the operating conditions change, they do not change very frequently. 
For e.g., in traffic signal control, the conditions change when vehicular inflow rate changes. 
It is observed that the vehicular rate remains constant for at least three to four hours in a single day before changing, depending on rush-hour traffic and 
early morning or late night traffic. 
Hence before every change, we get sufficient number of state and reward samples for our method. Additionally, in applications, the pattern of context changes is also usually known.

%% file: expres-sensor.tex
We consider the model described in \cite{prabuchandran2013q} which proposes an energy management (EM) MDP model for a sensor node with energy harvesting capabilities. 
Sensor node has a finite energy buffer to store the energy harvested from the ambient environment and a finite data buffer to store the data generated. 
The authors assume that energy units are harvested at a mean rate of $\lambda_E$, 
while data bits are generated at a mean rate of $\lambda_D$.   
The state of the system comprises of the current levels of energy and data buffers and the RL agent needs to decide on the number of energy units to be used
for transmission. The actual number of data bits transmitted is a non-linear function of the number of energy units utilized.
The RL agent needs to minimize the long-term discounted cost by finding a suitable policy. The immediate cost per step is the queue length of data buffer after successful transmission. 
In \cite{prabuchandran2013q}, model information is unknown and hence QL is used to find optimal EM policies.
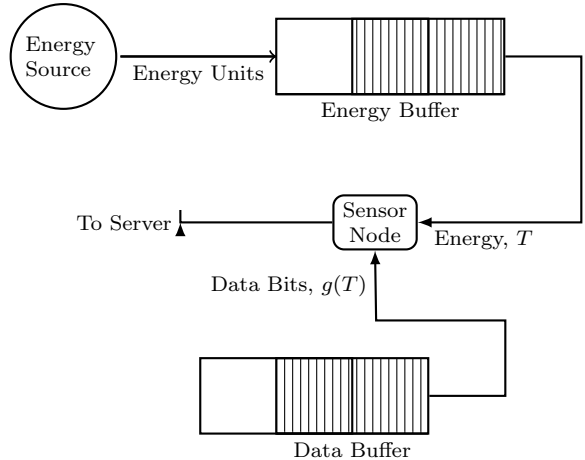
\begin{figure}[h!]
\begin{center}
\input{sensor-figure}
\end{center}
\caption{Energy harvesting model with a single sensor node with finite energy and data buffers} \label{fig:sensor}
\end{figure}

A sensor which is designed to harvest energy from the ambient environment, like for e.g., solar energy, has to appropriately modify its policy based on how $\lambda_E$ changes with day timings. We assume that the sensor monitors a physical system which generates data at a fixed rate that does not change over time.
A change in $\lambda_E$ gives rise to non-stationary environments. 
We consider this scenario and show that our method is effective in handling changing mean rate of energy harvest, when compared to QL, repeated update QL (RUQL) \cite{ruql}.
In our experiments, the exploration strategy used is 
$\epsilon$-greedy with $\epsilon = 0.1$. We analyze our method and QL, RUQL for regret when the environmental model changes once.
The number of iterations for learning phase is set to $4000$ with a changepoint at $2000$. 
\begin{table}[h]
\begin{center}

\begin{tabular}{|c|c|c|}
\hline
Method                     & Mean  $\pm$ SD             & Median        \\\hline
Context QL                 &  $498.75 \pm 48.78$        & $500$          \\\hline  
RUQL                       &  $803.7  \pm 121.3673$     & $800.5$        \\\hline
QL                         &  $675.5  \pm 50.96$        & $677$    \\\hline

\end{tabular}

\caption{Long-run discounted cost obtained by Context QL, RUQL and QL with $\epsilon$-greedy exploration.}
\label{table:sensor}
\end{center}
\end{table}
The long-run discounted cost obtained by our method, QL and RUQL is shown in Table \ref{table:sensor}. We are unable to compute the regret of these algorithms since model is unknown.

From Table \ref{table:sensor}, it is clear that Context QL does better compared to other model-free non-stationary algorithms like RUQL.
The purpose of evaluating policy learnt by QL in non-stationary environments is that if policy learnt by QL has performance which is comparable to Context QL, then it is easier to use QL albeit with a slight loss in performance. However, from Table \ref{table:sensor}, we observe that by using QL, RUQL, there is considerable loss in performance when environment contexts change. This is an implication of the fact that Context QL remembers policies for the environment models and updates the right set of Q values corresponding to the model. QL, RUQL on the other hand update the same set of Q values for the two different contexts.
This means the policy learnt by QL, RUQL is good only for the context which was the last to be observed in the sequence of contexts $M_{\theta_1}, M_{\theta_2},\ldots,M_{\theta_n}$, where $n = 2$ in the above experiments. For the same sequence of models, ContextQL learns a good policy for each of these contexts and stores them in the form of Q values. While evaluating, we detect the change in context and follow the appropriate policy for the next model in the sequence.

%% file: sensor-figure.tex
\begin{tikzpicture}[node distance = 14em, auto, thick]

    \node[draw,rectangle,minimum width=2cm,minimum height=1cm,text width=3em,text centered,pattern=vertical lines] at(1,-1) (B) {};
    \node[draw,rectangle,minimum width=2cm,minimum height=1cm,text width=3em,text centered] at(0,-1) (B1) {};
    \draw (0,-1.5) --node[midway,below]{Data Buffer} (2,-1.5);

    \node[draw,rectangle,rounded corners,text width=3em,text centered]  at (1.3,1.3) (sensor) {Sensor Node};

    \node[draw,rectangle,minimum width=2cm,minimum height=1cm,text width=3em,text centered,pattern=vertical lines] at(2,3.5) (E) {};
    \node[draw,rectangle,minimum width=2cm,minimum height=1cm,text width=3em,text centered] at(1,3.5) (E1) {};
    \draw (0,3) --node[midway,below]{Energy Buffer} (3,3);
    
    \node[draw,circle,minimum size=0.5cm,text width=1cm] at (-2.8,3.5) (esrc) {Energy Source};
    \draw[->] (-2.05,3.5) --++ (2.05,0);
    \draw (-2.05,3.5) --node[midway,below]{Energy Units} (0,3.5);
    
    \path [line] (E.east) -- ++(1,0) --++(0,-2.2) --++(-1,0) --node [near start,align=center] {Energy, $T$}(sensor.east);
    
     \path [line]  (B.east) --++(1,0) --++(0,1) --++(-1.7,0) --node [midway,align=center] {Data Bits, $g(T)$}(sensor.south);
     
     \path [line] (sensor.west) --++(-2,0) -- node[near end,align=center] {To Server} ++ (0,0);

    
    
\end{tikzpicture}

%% file: expres-traffic.tex
\begin{figure}[h]
\begin{center}
 \includegraphics[scale=0.3]{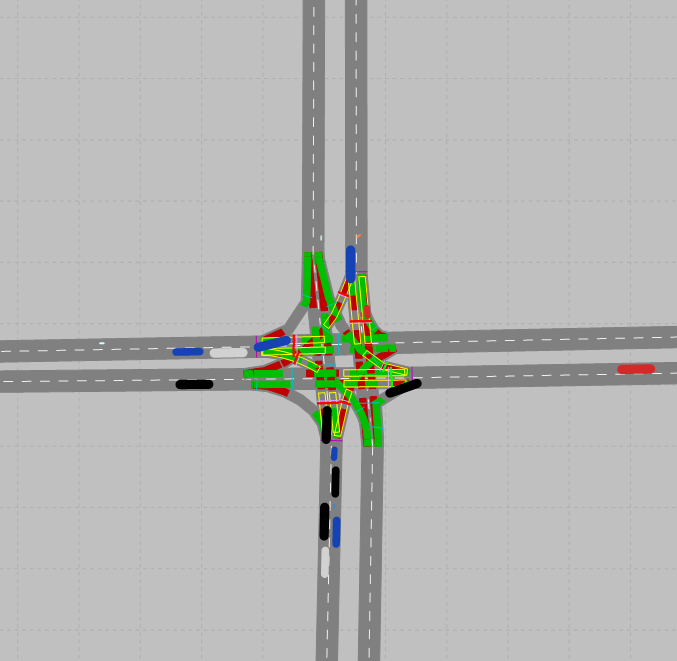}
 \caption{Illustration of the vehicular traffic junction - the junction has four incoming lanes, each having a lane distance of $150$ m. The green coloured areas at the junction indicate conflict zones, while the red coloured areas indicate reduced-speed zones. As seen, the vehicular input volume on the lanes is low.}
 \label{fig:traffic1}
 \end{center}
\end{figure}
\begin{figure}[h]
\begin{center}
 \includegraphics[scale=0.3]{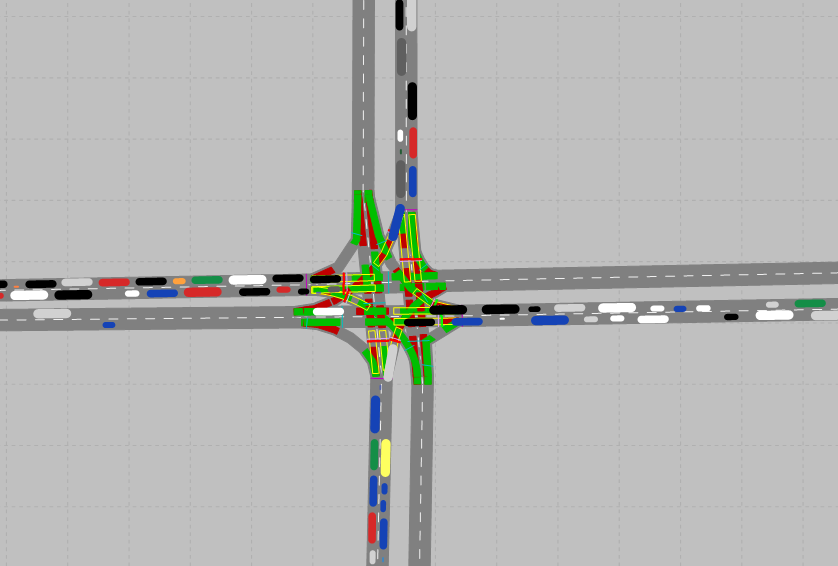}
 \caption{Illustration of the vehicular traffic junction when the vehicular input volume on the lanes is high. This scenario is observed when environment model dynamics ($P$) changes due to a change in vehicular input volumes.}
 \label{fig:traffic2}
 \end{center}
\end{figure}

As highlighted in Section \ref{sec:intro}, vehicular traffic signal control is a sequential decision-making problem. In the experiments,
we show that our method is effective in finding changes in vehicular patterns and learn the optimal policies for the same. 
The experimental setup consists of a single junction with four incoming lanes. 
The traffic junction is illustrated in Fig. \ref{fig:traffic1}, \ref{fig:traffic2} which are snapshots of the simulations carried out in VISSIM 
\footnote[1]{http://vision-traffic.ptvgroup.com/}. 
The junction is controlled by a signal. We model the traffic signal duration control as a MDP following \cite{prashla-tsc}. The state of the junction is the information consisting of queue lengths of all incoming lanes and the current phase. The phase indicates which incoming lane should be given the green signal. In order to tackle the state space dimensionality, we aggregate the queue lengths of lanes as $\textnormal{low}=0$, $\textnormal{medium} = 1$ and $\textnormal{high} = 2$. Hence, if a lane congestion level is one-third of its length, then we say that the aggregated state of that lane is $0$. If the lane congestion level is higher than one-third of the lane distance but lower than two-thirds the distance, then aggregated state of that lane is $1$. For high congestion levels, the aggregated state is $2$. With this lane queue length aggregation scheme, the state space dimensionality is reduced to $3^4 \times 4$. The actions for the signal controller 
correspond to the set of green signal durations $\{20,25,\ldots,70\}$ in seconds. 
The immediate cost is the sum of the lane queue lengths and the RL agent must minimize the long-term discounted cost.

The traffic RL agent learns a policy using QL. We train the traffic RL agent for $10^6$ simulation seconds with a change in vehicular input volumes after the simulation has run for half the time.  
A change in the input vehicular volumes causes a change in the environment dynamics. 
The long-run discounted costs obtained by QL and Context QL are shown in Table \ref{table:traffic}. 

\begin{table}[H]
\begin{center}

\begin{tabular}{|c|c|c|}
\hline
Method                     & Mean $\pm$ SD                  & Median        \\\hline
Context QL                 &  $1100.022 \pm 34$             & $1000$          \\\hline  
QL                         &  $1400.1 \pm 67$               & $1300$    \\\hline
\end{tabular}
\caption{Long-run discounted cost obtained by Context QL and QL with $\epsilon$-greedy exploration.}
\label{table:traffic}
\end{center}
\end{table}

%% file: main.bbl
\begin{thebibliography}{46}
\providecommand{\natexlab}[1]{#1}
\providecommand{\url}[1]{{#1}}
\providecommand{\urlprefix}{URL }
\expandafter\ifx\csname urlstyle\endcsname\relax
  \providecommand{\doi}[1]{DOI~\discretionary{}{}{}#1}\else
  \providecommand{\doi}{DOI~\discretionary{}{}{}\begingroup
  \urlstyle{rm}\Url}\fi
\providecommand{\eprint}[2][]{\url{#2}}

\bibitem[{Abdallah and Kaisers(2016)}]{ruql}
Abdallah S, Kaisers M (2016) {Addressing Environment Non-Stationarity by
  Repeating Q-learning Updates}. Journal of Machine Learning Research
  17(46):1--31

\bibitem[{Abounadi et~al.(2001)Abounadi, Bertsekas, and Borkar}]{rvqi}
Abounadi J, Bertsekas D, Borkar V (2001) {Learning Algorithms for Markov
  Decision Processes with Average Cost}. SIAM Journal on Control and
  Optimization 40(3):681--698, \doi{10.1137/S0363012999361974}

\bibitem[{Andrychowicz et~al.(2019)}]{openai2018learning}
Andrychowicz M, et~al. (2019) {Learning dexterous in-hand manipulation}. The
  International Journal of Robotics Research \doi{10.1177/0278364919887447}

\bibitem[{{Banerjee} et~al.(2017){Banerjee}, {Miao Liu}, and {How}}]{taposh1}
{Banerjee} T, {Miao Liu}, {How} JP (2017) Quickest change detection approach to
  optimal control in markov decision processes with model changes. In: 2017
  American Control Conference (ACC), pp 399--405,
  \doi{10.23919/ACC.2017.7962986}

\bibitem[{Bertsekas(2013)}]{BertB}
Bertsekas D (2013) Dynamic Programming and Optimal Control, vol~II, 4th edn.
  Athena Scientific, Belmont,MA

\bibitem[{Cano and Krawczyk(2019)}]{useless6}
Cano A, Krawczyk B (2019) Evolving rule-based classifiers with genetic
  programming on gpus for drifting data streams. Pattern Recognition 87:248 --
  268, \doi{https://doi.org/10.1016/j.patcog.2018.10.024}

\bibitem[{Choi et~al.(2000{\natexlab{a}})Choi, Yeung, and Zhang}]{choi2}
Choi SP, Yeung DY, Zhang NL (2000{\natexlab{a}}) Hidden-{M}ode {M}arkov
  {D}ecision {P}rocesses for {N}onstationary {S}equential {D}ecision {M}aking.
  In: Sequence Learning, Springer, pp 264--287

\bibitem[{Choi et~al.(2000{\natexlab{b}})Choi, Yeung, and Zhang}]{choi1}
Choi SPM, Yeung DY, Zhang NL (2000{\natexlab{b}}) {An Environment Model for
  Nonstationary Reinforcement Learning}. In: Solla SA, Leen TK, M\"{u}ller K
  (eds) Advances in Neural Information Processing Systems 12, MIT Press, pp
  987--993

\bibitem[{Cs\'{a}ji and Monostori(2008)}]{csaji}
Cs\'{a}ji BC, Monostori L (2008) {Value Function Based Reinforcement Learning
  in Changing Markovian Environments}. J Mach Learn Res 9:1679–1709

\bibitem[{Dick et~al.(2014)Dick, Gy\"{o}rgy, and Szepesv\'{a}ri}]{dick}
Dick T, Gy\"{o}rgy A, Szepesv\'{a}ri C (2014) {Online Learning in Markov
  Decision Processes with Changing Cost Sequences}. In: Proceedings of the 31st
  International Conference on International Conference on Machine Learning -
  Volume 32, JMLR.org, ICML’14, p I–512–I–520

\bibitem[{Ding et~al.(2019)Ding, Du, Zhao, Wang, and Jia}]{useless2}
Ding S, Du W, Zhao X, Wang L, Jia W (2019) A new asynchronous reinforcement
  learning algorithm based on improved parallel {PSO}. Appl Intell
  49(12):4211--4222, \doi{10.1007/s10489-019-01487-4},
  \urlprefix\url{https://doi.org/10.1007/s10489-019-01487-4}

\bibitem[{Everett and Roberts(2018)}]{mal}
Everett R, Roberts S (2018) {Learning against non-stationary agents with
  opponent modelling and deep reinforcement learning}. In: 2018 AAAI Spring
  Symposium Series

\bibitem[{Hadoux et~al.(2014)Hadoux, Beynier, and Weng}]{hadoux}
Hadoux E, Beynier A, Weng P (2014) {Sequential Decision-Making under
  Non-stationary Environments via Sequential Change-point Detection}. In:
  {Learning over Multiple Contexts (LMCE)}, Nancy, France

\bibitem[{Hallak et~al.(2015)Hallak, Castro, and Mannor}]{cmdp}
Hallak A, Castro DD, Mannor S (2015) {Contextual Markov Decision Processes}.
  In: Proceedings of the 12th European Workshop on Reinforcement Learning (EWRL
  2015)

\bibitem[{Harel et~al.(2014)Harel, Mannor, El-Yaniv, and
  Crammer}]{harel2014concept}
Harel M, Mannor S, El-Yaniv R, Crammer K (2014) {Concept Drift Detection
  Through Resampling}. In: International Conference on Machine Learning, pp
  1009--1017

\bibitem[{{Iwashita} and {Papa}(2019)}]{concept_drift_survey}
{Iwashita} AS, {Papa} JP (2019) {An Overview on Concept Drift Learning}. IEEE
  Access 7:1532--1547, \doi{10.1109/ACCESS.2018.2886026}

\bibitem[{Jaksch et~al.(2010)Jaksch, Ortner, and Auer}]{ucrl2}
Jaksch T, Ortner R, Auer P (2010) {Near-optimal regret bounds for reinforcement
  learning}. Journal of Machine Learning Research 11:1563--1600

\bibitem[{Kaplanis et~al.(2019)}]{crl1}
Kaplanis C, et~al. (2019) Policy consolidation for continual reinforcement
  learning. In: Proceedings of the 36th International Conference on Machine
  Learning, PMLR, vol~97, pp 3242--3251

\bibitem[{Kemker et~al.(2018)}]{cf}
Kemker R, et~al. (2018) Measuring catastrophic forgetting in neural networks.
  In: Thirty-second AAAI conference on artificial intelligence

\bibitem[{Kolomvatsos and Anagnostopoulos(2017)}]{useless1}
Kolomvatsos K, Anagnostopoulos C (2017) Reinforcement learning for predictive
  analytics in smart cities. In: Informatics, Multidisciplinary Digital
  Publishing Institute, vol~4, p~16

\bibitem[{Konda and Tsitsiklis(2003)}]{ac1}
Konda VR, Tsitsiklis JN (2003) {On Actor-Critic Algorithms}. SIAM Journal on
  Control and Optimization 42(4):1143--1166

\bibitem[{Krawczyk and Cano(2018)}]{useless5}
Krawczyk B, Cano A (2018) Online ensemble learning with abstaining classifiers
  for drifting and noisy data streams. Applied Soft Computing 68:677 -- 692,
  \doi{https://doi.org/10.1016/j.asoc.2017.12.008}

\bibitem[{Levin et~al.(2006)Levin, Peres, and Wilmer}]{mixperes}
Levin DA, Peres Y, Wilmer EL (2006) {Markov Chains and Mixing Times}. American
  Mathematical Society

\bibitem[{Liebman et~al.(2018)Liebman, Zavesky, and Stone}]{rlconceptdrift}
Liebman E, Zavesky E, Stone P (2018) {A Stitch in Time - Autonomous Model
  Management via Reinforcement Learning}. In: Proceedings of the 17th
  International Conference on Autonomous Agents and MultiAgent Systems,
  International Foundation for Autonomous Agents and Multiagent Systems, AAMAS
  ’18, p 990–998

\bibitem[{Matteson and James(2014)}]{ecp}
Matteson DS, James NA (2014) {A Nonparametric Approach for Multiple Change
  Point Analysis of Multivariate Data}. Journal of the American Statistical
  Association 109(505):334--345

\bibitem[{Minka(2000)}]{minka}
Minka T (2000) {Estimating a Dirichlet distribution}

\bibitem[{{Mohammadi} and {Al-Fuqaha}(2018)}]{useless3}
{Mohammadi} M, {Al-Fuqaha} A (2018) Enabling cognitive smart cities using big
  data and machine learning: Approaches and challenges. IEEE Communications
  Magazine 56(2):94--101, \doi{10.1109/MCOM.2018.1700298}

\bibitem[{Nagabandi et~al.(2018)}]{iclr2019}
Nagabandi A, et~al. (2018) {Learning to Adapt: Meta-Learning for Model-Based
  Control}. CoRR abs/1803.11347,
  \urlprefix\url{http://arxiv.org/abs/1803.11347}

\bibitem[{{Niroui} et~al.(2019){Niroui}, {Zhang}, {Kashino}, and
  {Nejat}}]{robotics2}
{Niroui} F, {Zhang} K, {Kashino} Z, {Nejat} G (2019) {Deep Reinforcement
  Learning Robot for Search and Rescue Applications: Exploration in Unknown
  Cluttered Environments}. IEEE Robotics and Automation Letters 4(2):610--617,
  \doi{10.1109/LRA.2019.2891991}

\bibitem[{Ortner et~al.(2019)Ortner, Gajane, and Auer}]{gajane2019paper1}
Ortner R, Gajane P, Auer P (2019) {Variational Regret Bounds for Reinforcement
  Learning}. In: Proceedings of the 35th Conference on Uncertainty in
  Artificial Intelligence

\bibitem[{Page(1954)}]{page}
Page ES (1954) {Continuous Inspection Schemes}. Biometrika 41(1/2):100--115

\bibitem[{Prabuchandran et~al.(2013)Prabuchandran, Meena, and
  Bhatnagar}]{prabuchandran2013q}
Prabuchandran KJ, Meena SK, Bhatnagar S (2013) Q-learning based energy
  management policies for a single sensor node with finite buffer. Wireless
  Communications Letters, IEEE 2(1):82--85,
  \doi{10.1109/WCL.2012.112012.120754}

\bibitem[{Prabuchandran et~al.(2019)Prabuchandran, Singh, Dayama, and
  Pandit}]{prabukj}
Prabuchandran KJ, Singh N, Dayama P, Pandit V (2019) {C}hange {P}oint
  {D}etection for {C}ompositional {M}ultivariate {D}ata. arXiv
  \eprint{1901.04935}

\bibitem[{{Prashanth} and {Bhatnagar}(2011)}]{prashla-tsc}
{Prashanth} LA, {Bhatnagar} S (2011) Reinforcement learning with average cost
  for adaptive control of traffic lights at intersections. In: 2011 14th
  International IEEE Conference on Intelligent Transportation Systems (ITSC),
  pp 1640--1645, \doi{10.1109/ITSC.2011.6082823}

\bibitem[{Puterman(2005)}]{puterman}
Puterman ML (2005) {Markov Decision Processes: Discrete Stochastic Dynamic
  Programming}, 2nd edn. John Wiley \& Sons, Inc., New York, NY, USA

\bibitem[{{Roveri}(2019)}]{tnnls}
{Roveri} M (2019) {Learning Discrete-Time Markov Chains Under Concept Drift}.
  IEEE Transactions on Neural Networks and Learning Systems 30(9):2570--2582,
  \doi{10.1109/TNNLS.2018.2886956}

\bibitem[{Salkham and Cahill(2010)}]{soilse}
Salkham A, Cahill V (2010) {Soilse: A decentralized approach to optimization of
  fluctuating urban traffic using Reinforcement Learning}. In: 13th
  International IEEE Conference on Intelligent Transportation Systems, pp
  531--538, \doi{10.1109/ITSC.2010.5625145}

\bibitem[{Shiryaev(1963)}]{shiryaev}
Shiryaev A (1963) {On Optimum Methods in Quickest Detection Problems}. Theory
  of Probability \& Its Applications 8(1):22--46

\bibitem[{da~Silva et~al.(2006)da~Silva, Basso, Bazzan, and Engel}]{rlcd}
da~Silva BC, Basso EW, Bazzan ALC, Engel PM (2006) {Dealing with Non-Stationary
  Environments Using Context Detection}. In: Proceedings of the 23rd
  International Conference on Machine Learning, Association for Computing
  Machinery, ICML ’06, p 217–224, \doi{10.1145/1143844.1143872}

\bibitem[{Sutton and Barto(2018)}]{sutton}
Sutton RS, Barto AG (2018) { Reinforcement Learning: An Introduction}, 2nd edn.
  MIT Press, Cambridge, MA, USA

\bibitem[{Sutton et~al.(1999)Sutton, McAllester, Singh, and Mansour}]{pg}
Sutton RS, McAllester D, Singh S, Mansour Y (1999) {Policy Gradient Methods for
  Reinforcement Learning with Function Approximation}. In: Proceedings of the
  12th International Conference on Neural Information Processing Systems, pp
  1057--1063

\bibitem[{Tatbul et~al.(2018)Tatbul, Lee, Zdonik, Alam, and
  Gottschlich}]{precrecall}
Tatbul N, Lee TJ, Zdonik S, Alam M, Gottschlich J (2018) {Precision and Recall
  for Time Series}. In: Advances in Neural Information Processing Systems, pp
  1920--1930

\bibitem[{{Tijsma} et~al.(2016){Tijsma}, {Drugan}, and {Wiering}}]{exp}
{Tijsma} AD, {Drugan} MM, {Wiering} MA (2016) Comparing exploration strategies
  for q-learning in random stochastic mazes. In: 2016 IEEE Symposium Series on
  Computational Intelligence (SSCI), pp 1--8, \doi{10.1109/SSCI.2016.7849366}

\bibitem[{Watkins and Dayan(1992)}]{ql}
Watkins CJ, Dayan P (1992) {Q-learning}. Machine learning 8(3-4):279--292

\bibitem[{{Yu} and {Mannor}(2009)}]{smannor}
{Yu} JY, {Mannor} S (2009) Online learning in markov decision processes with
  arbitrarily changing rewards and transitions. In: 2009 International
  Conference on Game Theory for Networks, pp 314--322,
  \doi{10.1109/GAMENETS.2009.5137416}

\bibitem[{Zhao et~al.(2019)}]{useless4}
Zhao X, et~al. (2019) {Applications of Asynchronous Deep Reinforcement Learning
  Based on Dynamic Updating Weights}. Applied Intelligence 49(2):581–591,
  \doi{10.1007/s10489-018-1296-x},
  \urlprefix\url{https://doi.org/10.1007/s10489-018-1296-x}

\end{thebibliography}
